\documentclass[conference]{IEEEtran}
\IEEEoverridecommandlockouts
\usepackage{cite}
\usepackage{amsmath,amssymb,amsfonts}
\usepackage{algorithmic}
\usepackage{graphicx}
\usepackage{textcomp}
\usepackage{xcolor}
\def\BibTeX{{\rm B\kern-.05em{\sc i\kern-.025em b}\kern-.08em
    T\kern-.1667em\lower.7ex\hbox{E}\kern-.125emX}}
\begin{document}

\title{The Accuracy-Efficiency Paradox: Quantifying Net Energy Loss in On-Device Energy Forecasting\\
}

\author{\IEEEauthorblockN{Jaeik Jeong}
\IEEEauthorblockA{\textit{Energy ICT Research Section, ETRI} \\
Daejeon, Republic of Korea \\
jaeik1210@etri.re.kr}
\and
\IEEEauthorblockN{Tai-Yeon Ku}
\IEEEauthorblockA{\textit{Energy ICT Research Section, ETRI} \\
Daejeon, Republic of Korea \\
kutai@etri.re.kr}
\and
\IEEEauthorblockN{Wan-Ki Park}
\IEEEauthorblockA{\textit{Energy ICT Research Section, ETRI} \\
Daejeon, Republic of Korea \\
wkpark@etri.re.kr}
}

\maketitle

\begin{abstract}
Energy forecasting aims to maximize accuracy to ensure energy efficiency by reducing energy waste, an objective that applies equally to on-device forecasting for mission-critical edge environments, including military systems. However, this paper identifies the Accuracy-Efficiency Paradox: high-precision energy forecasting models can ironically trigger a net energy deficit. This stems from both edge AI's inference energy consumption and battery aging. We propose a Total Cost of Ownership (TCO) framework for energy forecasting, designed to minimize net energy loss. This framework treats not only inference energy consumption but also battery aging as a unified form of energy loss, as degradation represents a physical dissipation of the system's future energy-carrying capacity. We demonstrate that in thermally sensitive edge environments, energy saved by the superior precision of complex architectures is often outweighed by the total energy lost through their high operational intensity.
\end{abstract}


\section{Introduction}
Accurate energy forecasting is a critical necessity for minimizing energy waste and ensuring system sustainability. As concerns over latency and privacy increasingly shift processing to the resource-constrained network edge, the necessity of performing on-device energy forecasting has become more pronounced. For example, in mission-critical systems such as military operational units, on-device AI ensures immediate response and data security in isolated environments, where latency and privacy are paramount for tactical success. While on-device forecasting research often incorporates model compression or lightweight architectures to meet hardware constraints, the primary optimization goal remains improving predictive precision, under the persistent assumption that higher energy forecasting accuracy inherently ensures superior energy efficiency by reducing energy waste \cite{subhan2025lightweight}.

However, this focus on accuracy gives rise to a critical Accuracy-Efficiency Paradox that challenges this very assumption. While high-complexity models reduce external energy forecasting errors, they impose a dual internal energy burden: significant inference energy consumption \cite{tu2023unveiling} and localized thermal stress that exponentially accelerates battery aging according to the Arrhenius Law \cite{kucinskis2022arrhenius}. Consequently, the energy saved through marginal forecasting accuracy gains can be entirely offset by the internal energy loss—comprising both inference energy consumption and battery aging. Despite this severity, most existing frameworks evaluate energy forecasting performance in isolation \cite{subhan2025lightweight}, failing to account for the operational costs incurred by the forecasting process itself. Furthermore, even studies on inference energy consumption and battery aging have largely proceeded in isolation \cite{tu2023unveiling,kucinskis2022arrhenius}, remaining disconnected from the evaluation of energy forecasting performance.

To reconcile this paradox, we propose an optimization framework based on a Total Cost of Ownership (TCO) metric. This approach translates both inference energy consumption and battery aging into a common metric of energy loss. We define battery aging not merely as hardware wear, but as an irreversible loss of embodied energy that reduces the device's total energy-carrying potential. Our main contributions are twofold: we propose a TCO objective function that unifies energy forecasting error, inference energy consumption, and battery aging into a single energy loss metric, and we mathematically define the thresholds where increased energy forecasting accuracy results in a net system energy loss.

\section{Methodology}\label{sec:method}
This section develops an evaluation framework that integrates Arrhenius-derived battery aging cost into a unified Total Cost of Ownership (TCO) metric to quantify the net energy efficiency of on-device energy forecasting.

\subsection{Derivation of Battery Aging Cost}
To quantify the impact on battery aging, we derive the relationship between inference energy and the acceleration factor (AF). The Arrhenius equation for the AF is given by:
\begin{equation}
    \text{AF} = \exp \left[ \frac{E_a}{k_B} \left( \frac{1}{T_{use}} - \frac{1}{T_{stress}} \right) \right],
\end{equation}
where $E_a$ is the activation energy, $k_B$ is the Boltzmann constant, $T_{use}$ is the nominal operating temperature, and $T_{stress}$ is the elevated junction temperature induced by the high operational intensity. Consequently, $\Delta T = T_{stress} - T_{use}$ represents the localized temperature rise due to the computational process.

\begin{figure*}[t]
\includegraphics[width=\linewidth]{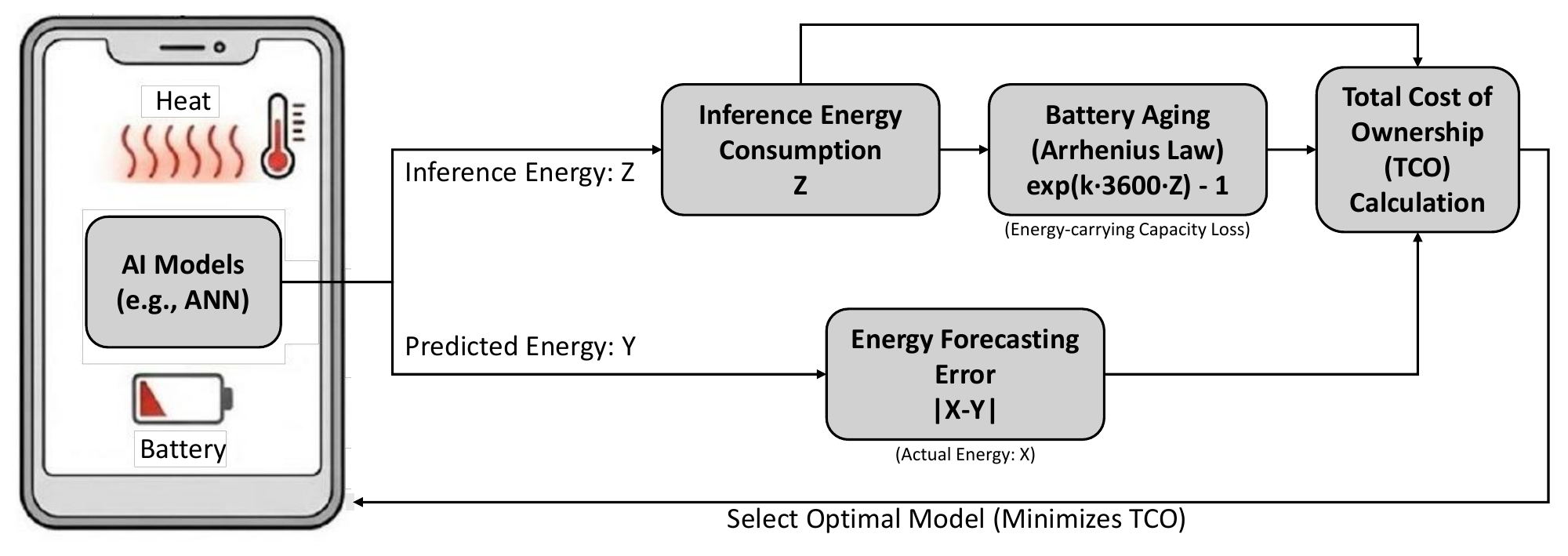}
 \centering
 \caption{TCO-based optimization framework: balancing energy forecasting error against inference energy consumption and battery aging.} 
 \label{fig:illustration}
\end{figure*}

In edge devices, operational energy ($Z$) is not converted into mechanical work but is dissipated as internal heat ($Q$). According to Landauer's principle \cite{chattopadhyay2025landauer}, the energy used for logic gate switching is fundamentally released as thermal energy. Given the compact, fanless form of embedded systems, this heat is transiently trapped within the device's thermal mass. According to the first law of thermodynamics yields $Q \approx 3600 \cdot Z$ (converting Wh to Joules). Incorporating the heat capacity $C$, the localized temperature increase $\Delta T$ is defined as:
\begin{equation}
\Delta T = \frac{Q}{C} \approx \frac{3600 \cdot Z}{C}.
\end{equation}
While adopting calories (cal) would scale the conversion constant to approximately 860, we strictly adhere to SI units (Joules) to ensure thermodynamic consistency with fundamental constants, such as the Boltzmann constant, used in the subsequent battery aging analysis. Substituting this Joule-based temperature increment into Eq. (1), we obtain:
\begin{equation}
\text{AF} = \exp \left[ \left( \frac{E_a}{k_B \cdot T_{use} \cdot T_{stress} \cdot C} \right) \cdot 3600 \cdot Z \right].
\end{equation}
Since both $T_{use}$ and $T_{stress}$ are absolute temperatures (Kelvin) and their variation is relatively small compared to their magnitude in typical operating ranges, the term $\frac{E_a}{k_B \cdot T_{use} \cdot T_{stress} \cdot C}$ can be treated as a consolidated environmental constant, $k$. Thus, the acceleration factor simplifies to $\text{AF} = \exp(k \cdot 3600 \cdot Z)$.

The AF is defined as the ratio of the stress-induced aging rate to the nominal rate. Let $r$ be the baseline aging rate at $T_{use}$ and $\Delta t_{\text{stress}}$ be the duration of elevated thermal stress. Then, the stress-induced aging during $\Delta t_{\text{stress}}$ is $\text{AF} \cdot r \cdot \Delta t_{\text{stress}}$, which makes the additional aging $(\text{AF} - 1) \cdot r \cdot \Delta t_{\text{stress}}$. The additional aging cost $\text{C}_{\text{aging}}$ is formulated as:
\begin{equation}
\text{C}_{\text{aging}}(Z) = c_{l} \cdot \left( e^{k \cdot 3600 \cdot Z} - 1 \right),
\end{equation}
where $c_{l}$ incorporates the unit cost of battery lifetime including $\Delta t_{\text{stress}}$. As inference energy $Z$ increases, the cost of battery aging escalates exponentially, particularly in thermally sensitive environments characterized by a high $k$ value and a quasi-adiabatic state that prolongs the thermal stress duration $\Delta t_{\text{stress}}$.

\subsection{Total Cost of Ownership (TCO) Formulation}
The proposed TCO metric at time $t$, $\text{TCO}_t$, is defined as:
\begin{equation}
\text{TCO}_t = c_{p}|X_t-Y_t| + c_{o}Z_t + c_{l} \left( e^{k \cdot 3600 \cdot Z_t} - 1 \right),
\end{equation}
where $X_t$ and $Y_t$ are actual and predicted energy, and $Z_t$ is the inference energy at time $t$. We introduce coefficients $c_{p}$ for forecasting error penalty, $c_{o}$ for operational cost, and $c_{l}$ for hardware lifetime cost. Crucially, this framework unifies disparate costs into a single dimension of energy inefficiency. Energy forecasting error ($|X_t - Y_t|$) represents system-level energy waste via suboptimal scheduling, while inference energy consumption ($Z_t$) is the direct energy waste of intelligence. Most importantly, the exponential battery aging term is treated not merely as a replacement cost, but as embodied energy waste, where degradation represents a permanent dissipation of the device’s future energy-carrying capacity.

As illustrated in Fig.~\ref{fig:illustration}, the TCO optimization engine prioritizes models that minimize this cumulative energy footprint rather than just forecasting error. By integrating real-time monitoring of inference energy consumption with Arrhenius-derived battery aging, the engine identifies the Accuracy-Efficiency Paradox where a nominally superior model becomes inferior due to its physical overhead. This closed-loop mechanism ensures that the deployed AI contributes to true net energy efficiency, preventing the shift of energy burdens from the software domain to irreversible hardware exhaustion.

\section{Experiment}\label{sec:experiment}
This section presents a numerical evaluation of the TCO framework using real-world data to demonstrate how environmental factors and operational intensity shift the optimal model choice.

\subsection{Dataset and Preprocessing}
This research used datasets from The Open AI Dataset Project (AI-Hub, South Korea), and all data information can be accessed through AI-Hub (www.aihub.or.kr). We utilized a high-resolution residential energy consumption dataset from Yeosu-si, spanning September 2021 to August 2022. The data consists of 8,760 hourly samples measured to three decimal places (kWh), providing the granularity necessary to detect marginal forecasting improvements. The dataset was chronologically partitioned into ten months for training, one month for validation, and one month for testing. During training, all values were normalized via min-max scaling to ensure stable gradient descent across architectures.

\subsection{Predictive Models and Training Strategy}To analyze the trade-off between accuracy and efficiency, we selected three models with varying complexity: Linear Regression (LR), Multi-Layer Perceptron (MLP), and Transformer. LR serves as the baseline, with weights analytically derived via ordinary least squares to ensure minimal overhead. The MLP represents a balanced artificial neural network (ANN), featuring a shallow architecture with two fully connected layers (64 hidden units) and ReLU activation. The Transformer represents high-complexity ANN architectures, utilizing a multi-head attention mechanism (4 heads) and two encoder layers with positional encoding. While the Transformer offers superior feature extraction, its self-attention mechanism significantly increases the computational load compared to the lighter alternatives.

ANN models (MLP and Transformer) were trained using the Adam optimizer with a learning rate ($0.0001$ to $0.001$) tuned via grid search. To ensure generalization, an early stopping strategy was employed based on the minimum Mean Absolute Error (MAE) on the validation set. All models utilize a sliding window approach, taking the preceding 24 hours of data as input for hour-ahead forecasting. This straightforward configuration is deliberately adopted to focus on validating the TCO Paradox through a controlled comparison of how varying model complexities respond to identical forecasting tasks under different thermal sensitivities.

\subsection{Experimental Results and Analysis}
The energy forecasting error and inference energy consumption of the three models are summarized in Table~\ref{tab:results}. Measurements were conducted in a representative on-device environment, with inference energy consumption estimated based on execution time at a constant operational power of 25W. While more granular energy tracking could be achieved through specialized tools like CodeCarbon in future studies \cite{anthony2020carbontracker}, this baseline is sufficient for validating the TCO paradox. As shown in Table~\ref{tab:results}, MLP and Transformer models achieved MAE gains of 1.7\% and 2.7\%, respectively, over the LR baseline. Although these gains appear numerically small—partly due to the strong linear autocorrelations in hour-ahead tasks where LR provides a near-optimal baseline—such improvements are recognized as significant contributions for enhancing system optimization. Notably, the inference energy consumption $Z$ remains several orders of magnitude smaller than the energy forecasting error $|X-Y|$. However, the critical concern is that the Transformer’s 2.7\% precision gain requires a disproportionate increase in $Z$, which acts as a catalyst for exponential battery aging.

\begin{table}[t]
\centering
\caption{Comparison of energy forecasting error ($|X-Y|$) and inference energy consumption ($Z$)}
\label{tab:results}
\begin{tabular}{lcc}
\hline
\textbf{Model} & \textbf{Average $|X-Y|$ (kWh)} & \textbf{Average $Z$ (Wh)} \\ \hline
LR          & 0.179833 & $0.1595 \times 10^{-6}$ \\
MLP         & 0.176825 & $0.2054 \times 10^{-6}$ \\
Transformer & 0.175035 & $6.4188 \times 10^{-6}$ \\ \hline
\end{tabular}
\end{table}

\begin{figure}[t]
    \centering
    \includegraphics[width=\columnwidth]{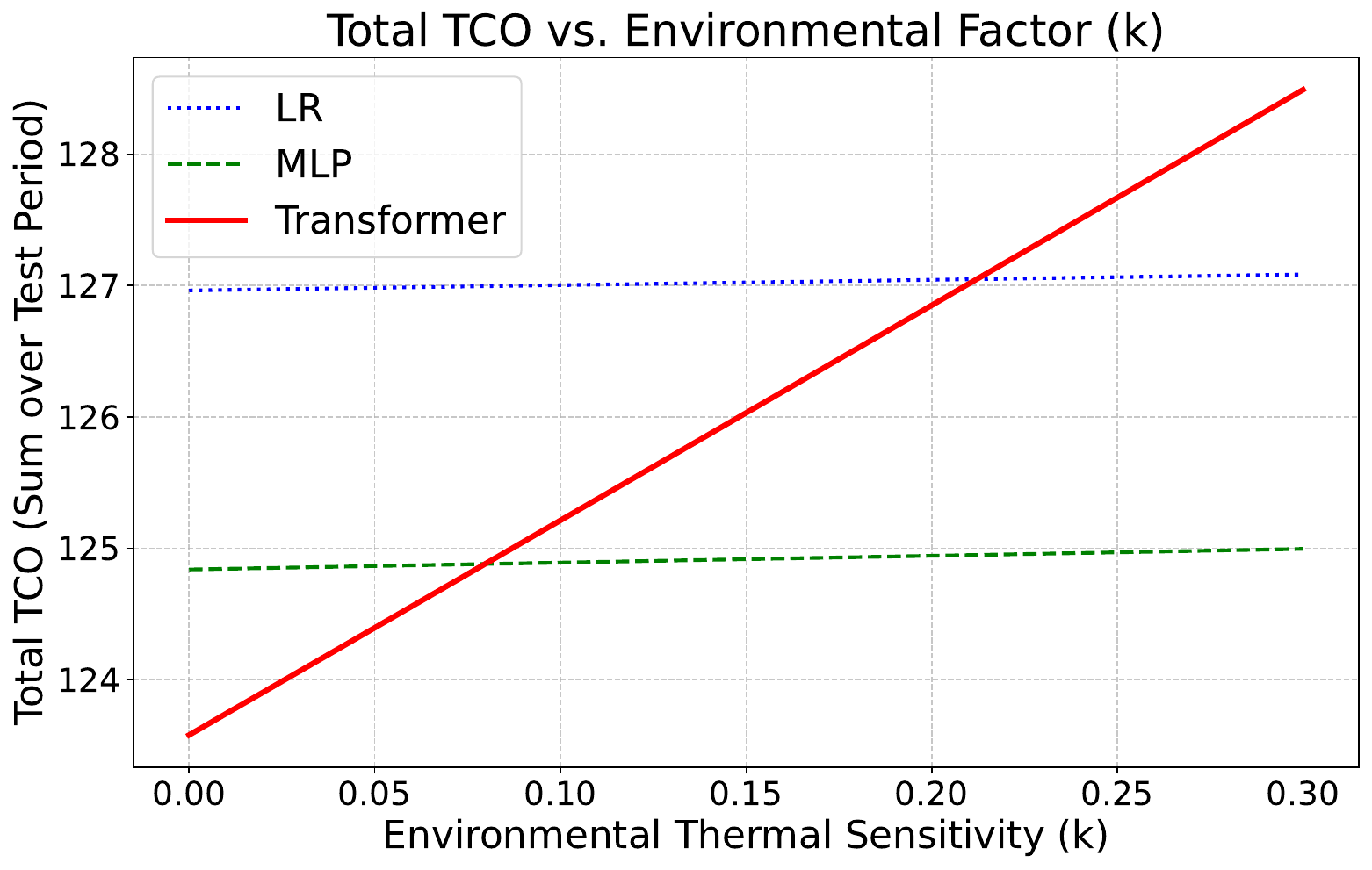}
    \caption{Total TCO vs. Environmental Factor $k$ (LR, MLP, and Transformer).}
    \label{fig:tco_k1}
    
    \vspace{0.5cm}
    
    \includegraphics[width=\columnwidth]{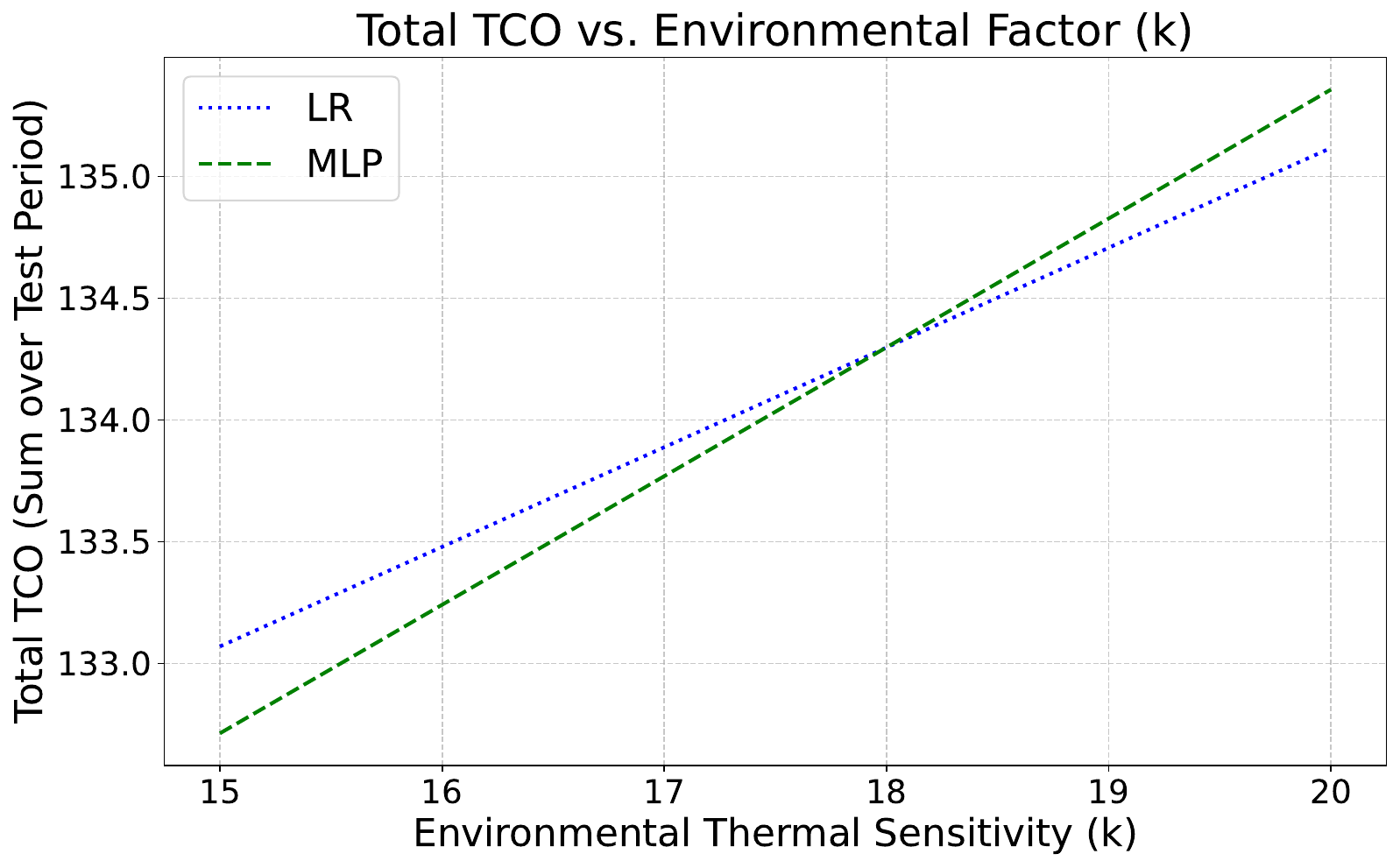}
    \caption{Total TCO vs. Environmental Factor $k$  (LR and MLP).}
    \label{fig:tco_k2}
\end{figure}

To validate the TCO paradox, we modulated the thermal sensitivity ($k$) while setting all cost coefficients ($c_p, c_o, c_l$) to 1. This setup allows for a transparent evaluation of how hardware degradation interacts with algorithmic precision under varying thermal constraints. As shown in Fig.~\ref{fig:tco_k1} and \ref{fig:tco_k2}, our analysis reveals critical trade-offs missed by traditional metrics. At low thermal sensitivity ($k < 0.08$), the Transformer achieves the lowest TCO as superior accuracy outweighs minimal battery aging costs. However, as $k$ increases, representing poorly cooled edge environments, the Transformer’s TCO escalates exponentially, becoming the least efficient choice for $k > 0.22$. Even the MLP exhibits a similar paradox against the LR baseline at extreme sensitivities ($k > 18.0$). These crossovers demonstrate that the optimal model is not static but dependent on the operating environment, where even negligible inference energy consumption differences can dictate the most sustainable choice.

\begin{figure}[t]
    \centering
    \includegraphics[width=\columnwidth]{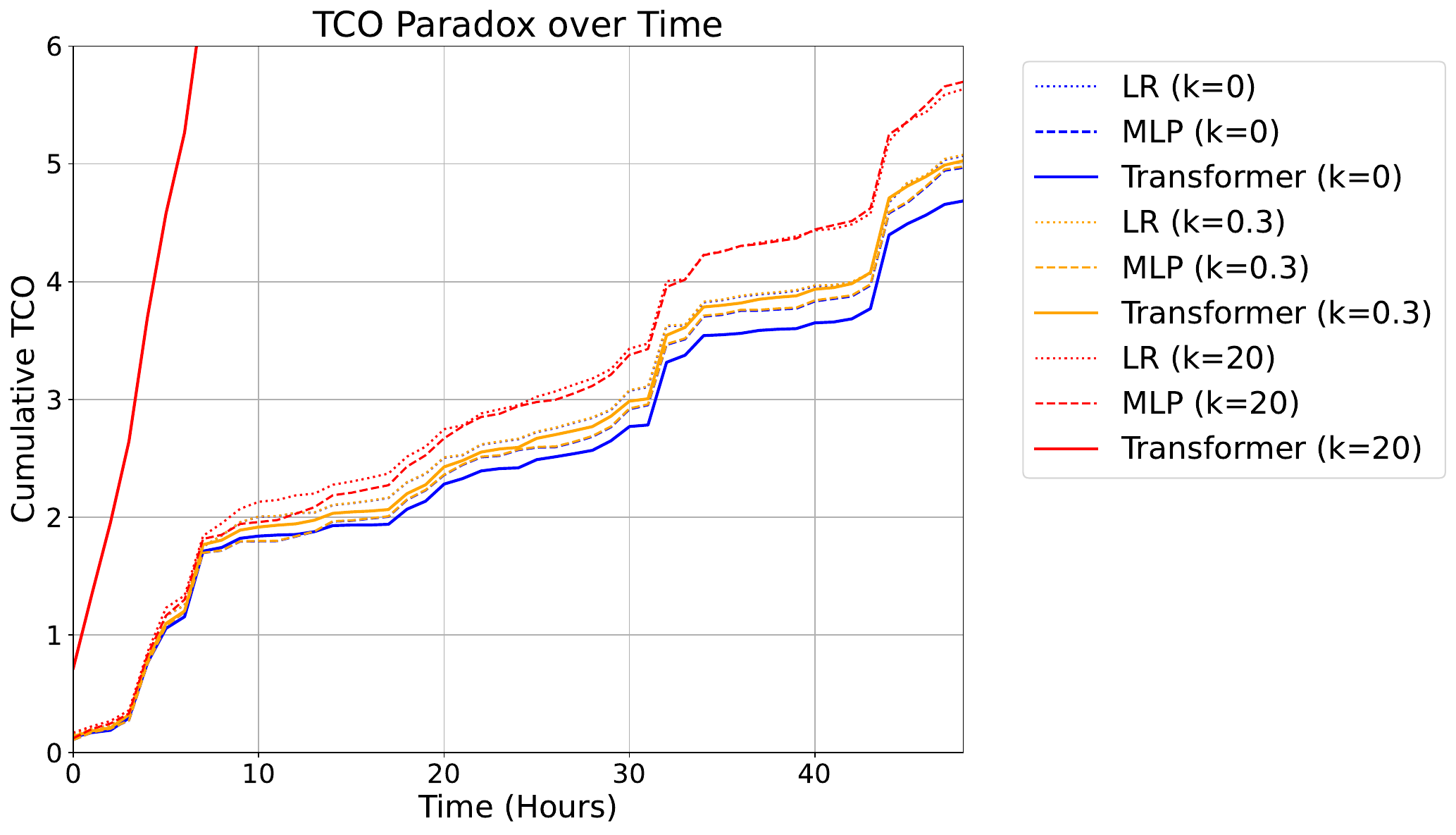}
    \caption{Dynamic accumulation of TCO over a 48-hour period under varying thermal sensitivity ($k$) conditions.}
    \label{fig:tco_time}
\end{figure}

Fig.~\ref{fig:tco_time} illustrates the 48-hour dynamic accumulation of TCO across different sensitivities. In stable environments ($k=0$ to $0.3$), the optimal choice shifts from Transformer to MLP as accuracy gains balance with rising battery aging costs. However, in high-sensitivity scenarios ($k=20$), the Transformer’s TCO escalates almost immediately, allowing the LR to eventually outperform even the MLP and emerge as the optimal solution with the minimum TCO. This real-time divergence confirms that as time progresses, physical degradation costs in thermally constrained environments can completely overwhelm superior predictive performance.

\section{Discussion}\label{sec:discussion}
The results highlight several research avenues. One might question whether a once-per-hour inference frequency could truly accelerate battery aging even for complex models; indeed, in an idle consumer device (e.g., a smartphone), rapid dissipation renders $\Delta t_{\text{stress}}$ negligible, making $c_l$ nearly zero. However, most edge devices operate under concurrent multi-purpose workloads \cite{woo2025exploring}. Adding high-complexity forecasting to a thermally saturated system creates a thermal bottleneck that hinders cooling, effectively inducing a quasi-adiabatic state, sustaining an elevated $\Delta t_{\text{stress}}$ (high $c_l$). Also, the overall analysis relies on simplified theoretical assumptions regarding thermal behavior, battery aging dynamics, and inference-to-heat conversion. Thus, empirically characterizing realistic values for $k$ and weighting coefficients ($c_p, c_o, c_l$) under diverse conditions remains a vital next step. 

The experimental evaluation is limited to simple hour-ahead residential energy forecasting tasks using a small set of baseline models. However, more complex domains such as multi-step-ahead or multi-site forecasting necessitate increasingly sophisticated ANN models, where the TCO Paradox must be rigorously examined. Furthermore, TCO components are context-dependent. In federated learning, operational energy consumption encompasses not only inference but also heavy local training \cite{wiesner2024fedzero}, which can escalate its impact to rival that of battery aging. In the case of data centers, operational energy consumption additionally encompasses significant cooling energy as well as inference and training. Also, data centers draw primary power from the grid, using Energy Storage Systems (ESS) in a supporting role to buffer surges \cite{rahman2026energy}. In these environments, ESS aging becomes less critical, especially with low-degradation technologies like Vanadium Redox Flow Batteries (VRFBs). Thus, TCO evaluation must be adaptable, as computation and longevity trade-offs shift from resource-constrained edge devices to hyperscale infrastructures.

\section{Conclusion}\label{sec:conclusion}
This paper introduced a TCO framework that integrates energy forecasting accuracy with inference energy consumption and battery aging to address the Accuracy-Efficiency Paradox in edge AI. Experimental results confirm that superior accuracy does not inherently equate to system-level efficiency when high operational intensity accelerates battery aging. Our analysis, grounded in the Arrhenius Law, reveals that marginal performance gains in high-complexity models are often offset by exponential battery aging costs in thermally constrained environments. By shifting from pure accuracy to a holistic TCO perspective, this study provides a vital tool for resolving the TCO paradox. Ultimately, our framework ensures the deployment of optimal, energy-aware edge models by balancing computational intelligence with physical reliability.

\section*{Acknowledgment}
This work was supported by the Korea Institute of Energy Technology Evaluation and Planning (KETEP) and the Ministry of Trade, Industry \& Energy (MOTIE) of the Republic of Korea. (No. RS-2025-02314925)

\bibliographystyle{IEEEtran}
\bibliography{ref}

\end{document}